\documentclass{article}

\PassOptionsToPackage{numbers, compress}{natbib}

\usepackage[preprint]{neurips_2022}

\usepackage[utf8]{inputenc} 
\usepackage[T1]{fontenc}    
\usepackage{hyperref}       
\usepackage{url}            
\usepackage{booktabs}       
\usepackage{amsfonts}       
\usepackage{nicefrac}       
\usepackage{microtype}      
\usepackage{xcolor}         

\usepackage{graphicx}
\usepackage{amsmath}
\usepackage{amsfonts,amssymb}
\usepackage{verbatim}

\title{YOLOSA: Object detection based on 2D local feature superimposed self-attention}

\author{%
  Weisheng Li\\
  Chongqing University of Posts and Telecommunications\\
  \texttt{liws@cqupt.edu.cn} \\
  \And
  Lin Huang  \\
  Chongqing University of Posts and Telecommunications \\
  \texttt{h72001346@163.com} \\
}

\begin{document}

\maketitle

\begin{abstract}

We analyzed the network structure of real-time object detection models and found that the features in the feature concatenation stage are very rich. Applying an attention module here can effectively improve the detection accuracy of the model. However, the commonly used attention module or self-attention module shows poor performance in detection accuracy and inference efficiency. Therefore, we propose a novel self-attention module, called 2D local feature superimposed self-attention, for the feature concatenation stage of the neck network. This self-attention module reflects global features through local features and local receptive fields. We also propose and optimize an efficient decoupled head and AB-OTA, and achieve SOTA results. Average precisions of 49.0\% (71FPS, 14ms), 46.1\% (85FPS, 11.7ms), and 39.1\% (107FPS, 9.3ms) were obtained for large, medium, and small-scale models built using our proposed improvements. Our models exceeded YOLOv5 by 0.8\% -- 3.1\% in average precision.
\end{abstract}

\section{Introduction}

\label{intro}
In recent years, the application scenarios of real-time object detection\cite{yolo1,yolo2,yolo3,yolo4,syolo4,yolo5,yolopp,yolopp2,yolof,yolox,focal,efficientdet,asff,ssd} are more and more diverse, including safety protection, defect detection, and visual positioning. The application of object detection greatly reduces the labor cost and improves production efficiency, but it will also lead to the reduction of employment opportunities and the unemployment of some people. However, after some practical investigations of some enterprises in the industrial field, the problem faced by many enterprises is not the problem of labor cost, but the problem of difficult recruitment and the high cost of employee induction training. Therefore, the application of object detection in some industries has a very positive effect, which can help enterprises reduce the recruitment risk and reduce the entry threshold of employees. In addition, the object detection algorithm has its own detection limitations. In many practical application scenarios, man-machine collaboration is better than relying entirely on machines and algorithms. Manual re-inspection is very important, which will increase employment opportunities and solve the problem of unemployment.

Attention mechanisms\cite{senet,cbam,selfattention} have been widely used in vision tasks, such as SENet\cite{senet} and CBAM\cite{cbam}. However, they have seldom been applied for YOLO series\cite{yolo1,yolo2,yolo3,yolo4,syolo4,yolo5,yolopp,yolopp2,yolof,yolox} object detection. One such example is the spatial attention mechanism\cite{cbam} being used in YOLOv4\cite{yolo4}. Along with ViT\cite{vit}, self-attention\cite{selfattention} and multi-head self-attention\cite{selfattention} are increasingly being adopted in the field of computer vision, and the advantages of the global attention mechanism\cite{selfattention} are becoming increasingly obvious. However, one-stage object detection\cite{yolo1,yolo2,yolo3,yolo4,syolo4,yolo5,yolopp,yolopp2,yolof,yolox,focal,efficientdet,asff,ssd} requires a very high inference speed. Owing to a large number of calculations and parameters, the application of the self-attention module substantially affects the speed of one-stage object detection. Therefore, we propose a self-attention module more suitable for one-stage object detection while ensuring that the inference speed of the original algorithm is retained.

\begin{figure}
\centering
\includegraphics[width=0.8\textwidth]{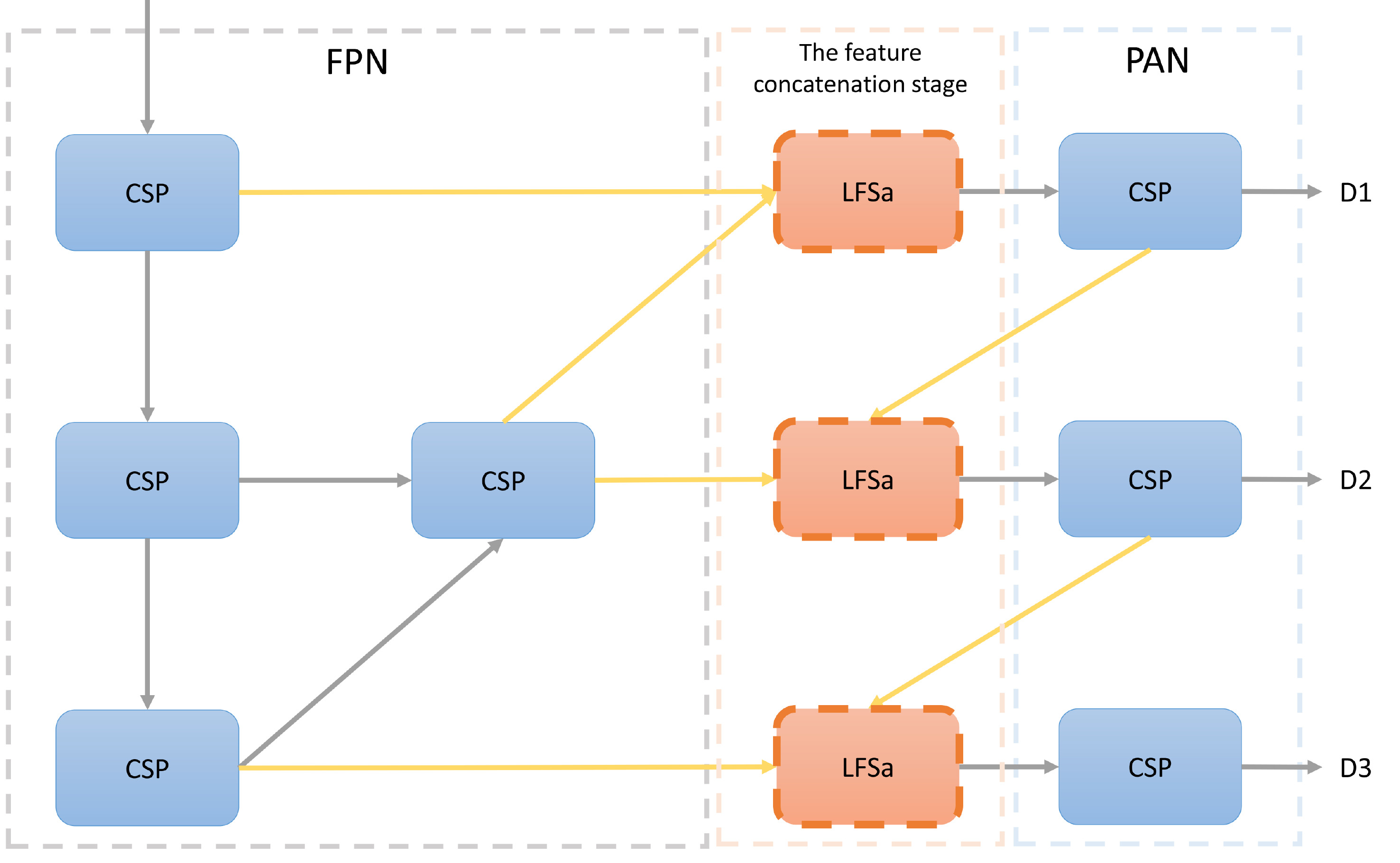}
\caption{Position of LFSa module applied in the FPN\cite{fpn} + PAN\cite{pan} structure as well as feature concatenation stage.}
\label{fig: pos}
\end{figure}

In object detection, an attention module is mainly applied in two places: in the backbone network and in the head network\cite{yolo4}. As the feature information contained in the feature concatenation stage (Fig.~\ref{fig: pos}), before the object detection feature is input into the detection head, is very rich, there will be features from various levels, and adding attention here is very conducive to improving the accuracy of object detection. We tried a variety of attention modules (including CAM\cite{senet} and CBAM\cite{cbam}), but they did not substantially improve the accuracy. Therefore, we turned our attention to self-attention. However, given the large number of channels and the large size of the feature map, using the multi-head self-attention module directly will lead to extensive network calculation and low network inference speed. To overcome this issue, we propose a novel self-attention module called 2D local feature superposition self-attention. We also add an efficient decoupled head and AB-OTA to the model. These two approaches not only optimize the inference efficiency of the anchor-based model but also substantially reduce the number of calculations and parameters of the model.

\begin{figure}
\centering
\includegraphics[width=0.8\textwidth]{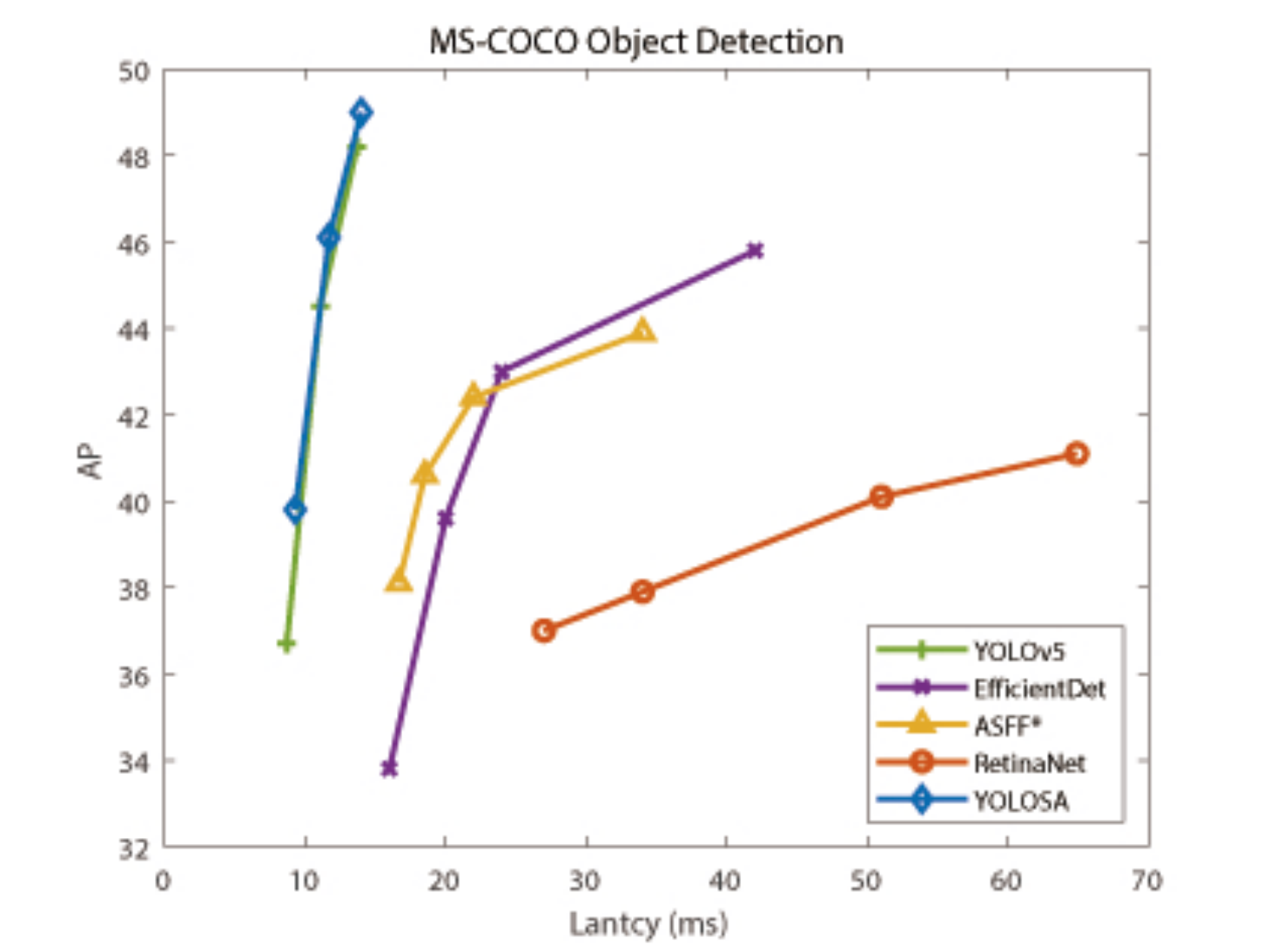}
\caption{ Comparison of AP and FPS(RTX3090) of the proposed YOLOSA and other SOTA object detectors.}
\label{fig: data}
\end{figure}

\textbf{Contribution}. The average precision (AP) of our model showed an improvement of 0.8\%--3.1\% compared to YOLOv5, where the large-scale model had an AP of 49.0\% (+0.8\%), the medium-scale model had an AP of 46.1\% (+1.5\%), and the small-scale model had an AP of 39.1\% (+3.1\%) (Fig.~\ref{fig: data}). The contributions of our paper can be summarized as follows:

\begin{enumerate} 
\item We proposed the 2D local feature superposition self-attention module. This is an efficient module that can directly reflect the global weighted features from the local weighted features superimposed by the weighted feature map of the rows and columns.

\item We added an optimized decoupled head\cite{yolox}, called efficient decoupled head, to the model, which considerably improved the inference efficiency of the model.

\item Based on SIM-OTA\cite{yolox}, we proposed the AB-OTA for the anchor-based model, which is used to select the best ground truth in the same anchor and grid in the training process.
\end{enumerate}

\section{YOLOSA}

\subsection{2D local feature superimposed self-attention (LFSa)}

\begin{figure}
\centering
\includegraphics[width=1\textwidth]{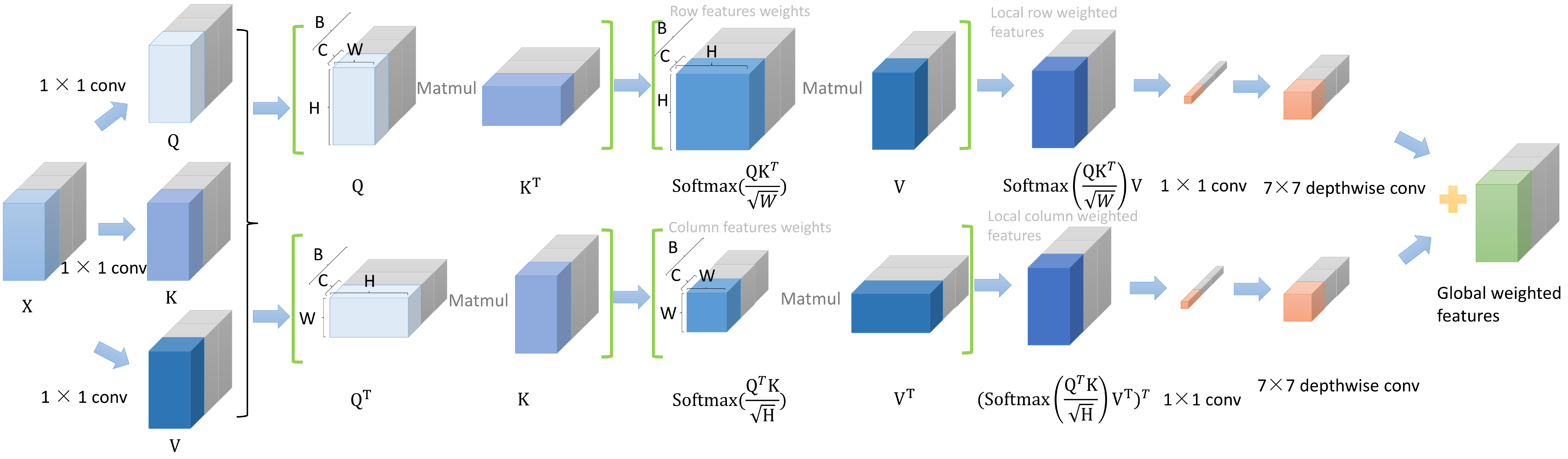}
\caption{The figure shows the calculation process of 2D local feature superimposed self-attention (LFSa).The green bracket part is the matmul product of the matrix, which aims to calculate the feature weights and weighted features of the rows or columns of the feature maps. Then, after 1 * 1 convolution and 7 * 7 depthwise convolution, the row or column weighted feature maps is added to obtain the pixel based global weighted feature maps. }
\label{fig:lfs}
\end{figure}

In the neck structure\cite{yolo4} of real-time object detection\cite{yolo1,yolo2,yolo3,yolo4,syolo4,yolo5,yolopp,yolopp2,yolof,yolox,focal,efficientdet,asff,ssd}, the most popular approach is to use the FPN\cite{fpn} + PAN\cite{pan} structure. The feature concatenation stage of this structure (before input to the detection head) concatenates the output of the feature maps from multiple levels of the backbone network. Therefore, the feature information gathered here plays a vital role in the object inference of the detection head. We focused on adding an attention module at this stage and hoped to improve the model’s accuracy performance without affecting its inference speed. After adding the CAM\cite{senet} or CBAM\cite{cbam} at this stage, we found that the accuracy of the model did not improve much (Table ~\ref{table: attn}). We also tried to add the multi-head self-attention\cite{selfattention} module here, but owing to the sharp increase in the number of model calculations and parameters (The matrix with the shape of $channels \times (h\cdot w)$ is multiplied by its own transpose), the inference speed of the model is very low, which cannot meet the requirements of real-time object detection. To overcome these challenges, we propose a novel self-attention module called 2D local feature superimposed self-attention (Fig.~\ref{fig:lfs}).

\begin{figure}
\centering
\includegraphics[width=1\textwidth]{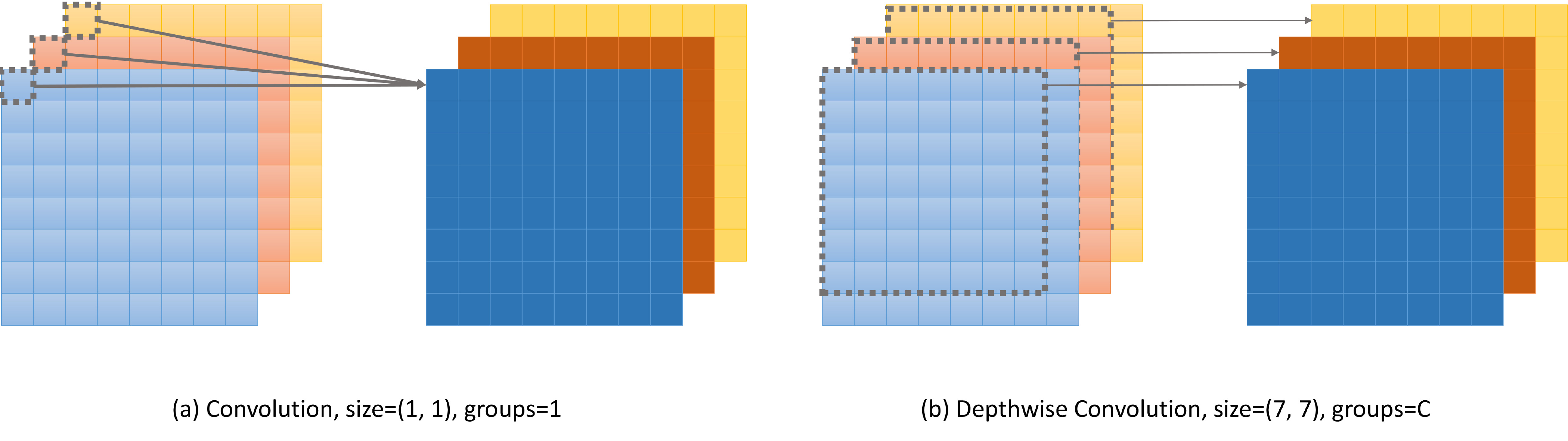}
\caption{Figure (a) shows the convolution of size = 1 and groups = 1, which is used to get through the global weighted features of rows or columns in the single channel feature map with the channel dimension in LFSa. Figure (b) shows the depthwise convolution\cite{mobilenets} of size = 7 and groups = C, which is used to increase the receptive field of sensing the feature map in each channel.}
\label{fig:lfs_conv}
\end{figure}

The calculation procedure of LFSa is very similar to that of the self-attention module. We did not calculate the global feature weights for the feature maps of all channels, because doing so would entail a large number of calculations. We only calculated the local feature weights for rows or columns in the feature map of each channel. Assume $X\in\mathbb{R}^{C \times H \times W}$ is the input feature map, and $Q$, $K$, and $V$ are the feature maps of $X$ after convolution with the convolving kernel size of $1\times1$. $Q_i$, $K_i$, and $V_i$ represent the feature map of one channel $i \in \{0,1,2,...,C\}$. The calculation process can be summarized as follows:

\begin{align}
F_i^{row}(X) &= Softmax(\frac{Q_{i}(X)K_{i}^{T}(X)}{\sqrt{W}}) V_i(X) \label{eq:lfs1} \\
F_i^{col}(X) &= (Softmax(\frac{Q_{i}^{T}(X)K_{i}(X)}{\sqrt{H}})V_i^T(X))^T  \label{eq:lfs2} 
\end{align}

where $F_i^{row}(X)$, and $F_i^{col}(X)$ are globally weighted features for the rows or columns of one channel feature map. Relative to the feature map of one channel, this weighted feature is global, but relative to all channels of one picture, it is local. Next, we used the convolution with a convolving kernel size of $1\times1$ to extract the channel features of  $F^{row}(X)$, and $F^{col}(X)$ in the dimension of the channel (Fig.~\ref{fig:lfs_conv} (a)). As the feature map of each channel is the global weighted feature of the row or column, it is equivalent to using the local receptive field to obtain the global weighted feature of the entire picture based on the row and column in the dimension of the channel. Therefore, the perception of global features is completed through these two steps of extracting local features. This method substantially reduces the amount of calculation and parameters of the model.

As we use $1\times1$ convolution to make the receptive field of the feature map of each channel relatively small, if we increase the size of the convolving kernel, it will lead to an increase in the number of calculations and parameters. Therefore, we added a depthwise convolution\cite{mobilenets} (Fig.~\ref{fig:lfs_conv} (b)) after the $1\times1$ convolution and used the $7\times7$ large-size convolving kernel to increase the perception of the feature map information of each channel. As each convolving kernel is independent of each channel, the increase of calculation and parameters is within an acceptable range while increasing the receptive field.

In the last step, we added the row global weighted feature maps and the column global weighted feature maps. In this manner, the globally weighted feature map of pixels based on all channel feature maps is obtained. The idea of the global weighted of the self-attention model is realized in another manner, which reduces the number of calculations and parameters of the model substantially. The calculation process can be summarized as follows:

\begin{align}
LFSa(X) &= X + DepthwiseConv(Conv(F^{row}(X))) + DepthwiseConv(Conv(F^{col}(X))) \label{eq:lfs3}
\end{align}

where $Conv$ represents the convolution layer with a convolving kernel size of $1\times1$ and groups of 1, and $DepthwiseConv$ represents the depthwise-convolution layer with a convolving kernel size of $7\times7$ and groups equal to the number of channels.

\begin{figure}
\centering
\includegraphics[width=1\textwidth]{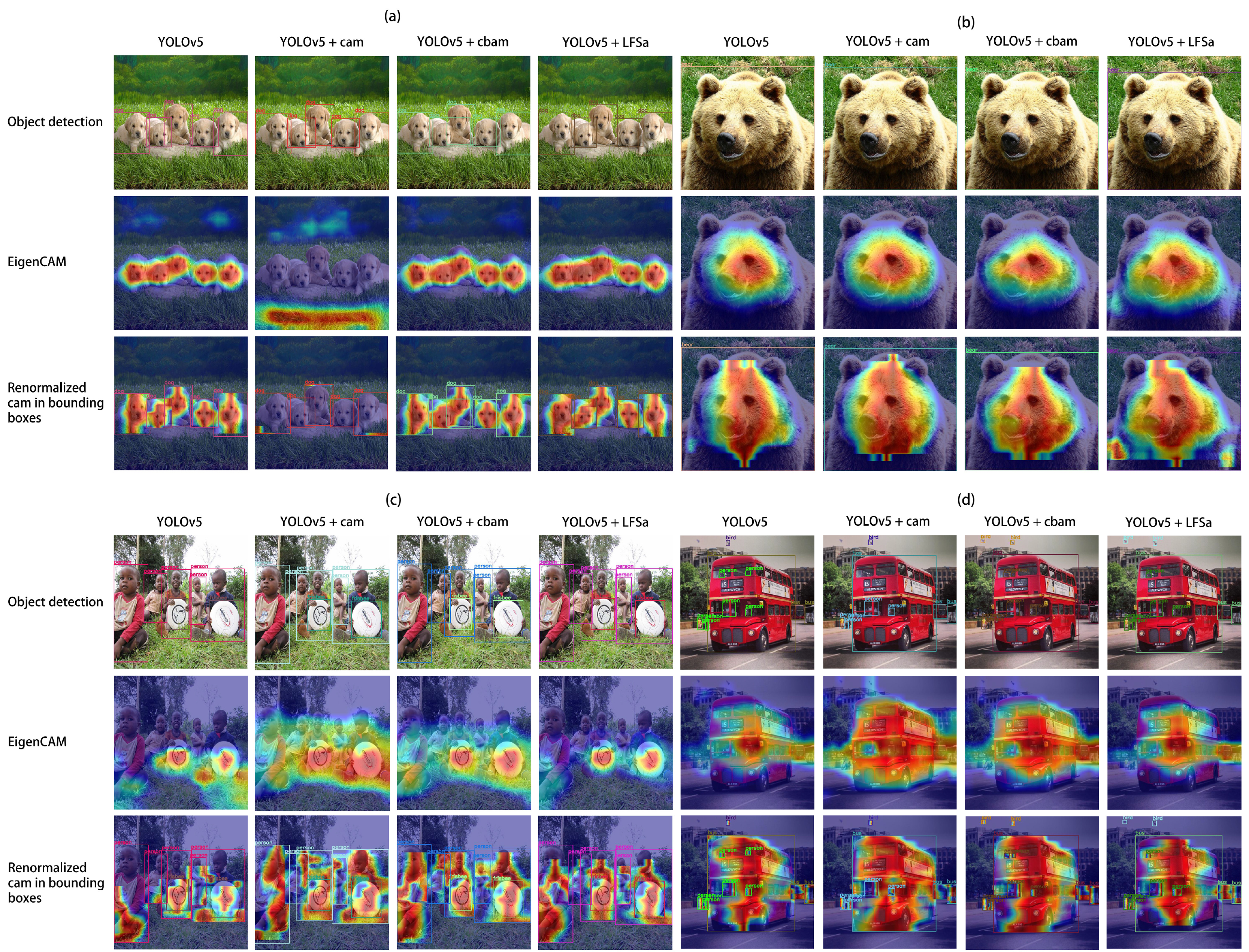}
\caption{In the figure, the first row contains the object detection results, the second row shows the visualization results of Eigen-CAM\cite{eigen}, and the third row shows the results after removing the heatmap data outside the bounding box and scaling the heatmap in each bounding box.}
\label{fig:lfs_pic}
\end{figure}

To more intuitively compare the advantages of our LFSa with other attention modules, we used the Eigen-CAM\cite{eigen} visualization results to analyze the focus of the network on the picture (Fig.~\ref{fig:lfs_pic}). Eigen-CAM computes and visualizes the principle components of the learned features/representations from the convolutional layers. Thus, we can intuitively determine the attention of various modules to the detection object through the heatmap in the picture and bounding box. In fig.~\ref{fig:lfs_pic}, our LFSa module focuses more on the object and maximizes the range of attention within the bounding box of the object. In addition, our model pays less attention to the area outside the bounding box of the object, i.e., the background has less interference with the object. Our module also globally stratifies the attention of multiple objects. In Fig.~\ref{fig:lfs_pic} (a), we can see that YOLOv5\cite{yolo5}+CAM\cite{senet}’s attention range has a very large deviation. In addition, if we observe the image in the second row, we can find that except for our LFSa module, the backgrounds of other model inference images have spots of various sizes on the top of the object. This shows that the background has noticeable interference with the detected object, except for in our module. In Fig.~\ref{fig:lfs_pic} (b), compared with other models, our LFSa module’s attention range is closer to the edge of the object and pays more attention to the details. In Fig.~\ref{fig:lfs_pic} (c), our LFSa module divides the attention of different objects from a global perspective, so that the attention of different objects is obviously stratified, and the attention of the same object is roughly the same. This stratification is more obvious in the attention of multiple objects shown in Fig.~\ref{fig:lfs_pic} (d). Therefore, we can see from Fig.~\ref{fig:lfs_pic} that our method is superior to other methods.

We also compared the AP of the LFSa module with other attention modules based on the YOLOv5l\cite{yolo5} model. As presented in Table ~\ref{table: attn}, our LFSa module achieved very good performance, with an increase of 0.7\% over the baseline on AP, and the growth rate is much higher than CAM\cite{senet} and CBAM\cite{cbam} modules.

\setlength{\tabcolsep}{4pt}
\begin{table}
\begin{center}
\caption{Comparison with different attention modules on MS-COCO val2017\cite{mscoco}}
\label{table: attn}
\begin{tabular}{lllllll}
\hline\noalign{\smallskip}
Models & AP(\%) & $AP_{50}$ & $AP_{75}$ & $AP_S$ & $AP_M$ & $AP_L$ \\
\noalign{\smallskip}
\hline
\noalign{\smallskip}
YOLOv5l & 48.2 & 66.9 & - & - & - & - \\
YOLOv5l+cam & 48.3 (\textcolor{green}{+0.1}) & 66.8 & 52.7 & 31.3 & 53.2 & 61.7 \\
YOLOv5l+cbam & 48.5 (\textcolor{green}{+0.3}) & 67.0 & 53.1 & 32.0 & 53.2 & 62.4 \\  
YOLOv5l+LFSa & 48.9 (\textcolor{green}{+0.7}) & 67.1 & 53.0 & 31.8 & 53.9 & 62.6 \\
\hline
\end{tabular}
\end{center}
\end{table}
\setlength{\tabcolsep}{1.4pt}

In general, our LFSa module combines the advantages of the self-attention module\cite{selfattention} and convolutional network. We first calculate the weighted features based on rows/columns of the feature map based on each channel in each picture and then connect the entire picture from the dimension of the channel through the local receptive field of the convolution network, which is the global feature based on the picture. This realizes the idea of the self-attention module from a more efficient perspective.


\subsection{Efficient Decoupled Head (EDH)}

\begin{figure}
\centering
\includegraphics[width=1\textwidth]{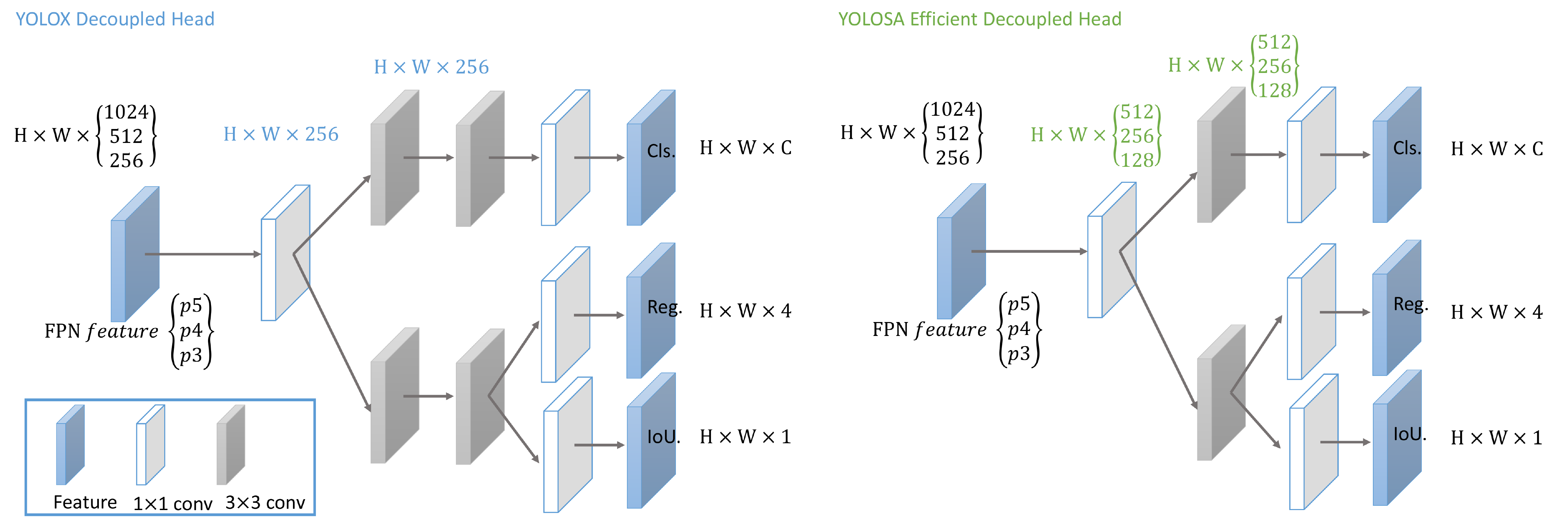}
\caption{The blue part is the number of network channels of YOLOX\cite{yolox}, while the green part is the number of network channels of the efficient decoupled head.}
\label{fig: edh}
\end{figure}

To resolve the conflict between classification and regression tasks, the decoupled head\cite{yolox} method is widely used in one-stage object detection\cite{yolo1,yolo2,yolo3,yolo4,syolo4,yolo5,yolopp,yolopp2,yolof,yolox,focal,efficientdet,asff,ssd} and two-stage object detection\cite{maskrcnn,cascadercnn,rfh,fastrcnn,fasterrcnn,rfcn,lrcnn}. YOLOX\cite{yolox} took the lead in applying the decoupled head\cite{yolox} method to YOLO series\cite{yolo1,yolo2,yolo3,yolo4,syolo4,yolo5,yolopp,yolopp2,yolof,yolox} models and obtained SOTA performance. However, when we applied the decoupled head method to the YOLOv5\cite{yolo5} model, the number of model calculation and the number of model parameters are considerably increased. This method has a noteworthy impact on the inference speed of our model. Accordingly, we made some improvements to the original method to reduce the amount of calculation in the model and improve the inference speed (Fig.~\ref{fig: edh}).

YOLOX\cite{yolox} passed the output of the FPN\cite{fpn} + PAN\cite{pan} structure through a $1\times1$ convolution with 256 channels. Our method passes all outputs through a $1\times1$ convolution in which the number of channels is halved and one $3\times3$ convolution was removed. As presented in Table ~\ref{table: edh}, our method achieves a slightly lower AP than decoupled head of YOLOX\cite{yolox}, but the number of calculations and parameters of the model are considerably reduced, which speeds up the inference speed of the model substantially.

\setlength{\tabcolsep}{4pt}
\begin{table}
\begin{center}
\caption{Comparison of Parameters, GFLOPs and AP between YOLOX Decoupled Head\cite{yolox} and YOLOSA Efficient Decoupled Head on MS-COCO val2017\cite{mscoco}.}
\label{table: edh}
\begin{tabular}{lllll}
\hline\noalign{\smallskip}
Models & AP(\%) & $AP_{50}$ & Parameters(M) & GFLOPs \\
\noalign{\smallskip}
\hline
\noalign{\smallskip}
YOLOv5l & 48.2 & 66.9 & 47.0 & 115.4 \\
YOLOv5l + Dcoupled Head& 48.9(\textcolor{green}{+0.7}) & 67.4(\textcolor{green}{+0.5}) & 50.3(\textcolor{gray}{+3.3}) & 150.1(\textcolor{gray}{+34.7}) \\
YOLOv5l + Efficient Decoupled Head& 48.6(\textcolor{green}{+0.4}) & 67.2(\textcolor{gray}{+0.3}) & 50.1(\textcolor{green}{+3.1}) & 121.2(\textcolor{green}{+5.8}) \\
\hline
\end{tabular}
\end{center}
\end{table}
\setlength{\tabcolsep}{1.4pt}

\subsection{AB-OTA}

The AB-OTA we proposed is an improvement based on YOLOv5\cite{yolo5}’s label assignment rules\cite{la1,la2,la3,la4,la5}, mainly referring to YOLOX’s SimOTA\cite{yolox}. YOLOv5\cite{yolo5} is an anchor-based object detection model. If the same grid on the same anchor of the model matches multiple ground truths, the label assignment rule is usually the last ground truth assignment principle. This leads to random label assignment. Therefore, we proposed a new method for label assignment of the YOLOv5 model. The main calculation process involves calculating the CIoU\cite{ciou} of the multiple matched ground truth with the predicted value, finding the two largest CIoUs of the ground truth, calculating the cost, and then selecting the one with the lowest cost as the most matched ground truth.

Assume that the ground truth of multiple matches is denoted by $g_i$, and the predicted value is $p_j$. The calculation process can be summarized as follows:

\begin{align}
val_j, idx_j &= TopK(CIoU(g_i,p_j), k=2)  \\
V_j, I_j &= Min(L_{idx_j,j}^{cls} + \lambda L_{idx_j,j}^{reg}) 
\end{align}

where $\lambda$ is a balancing coefficient. $L_{idx_j,j}^{cls}$ and $L_{idx_j,j}^{reg}$ are the classification loss and CIoU loss respectively. $TopK$ denotes the values/indexes that select the two largest values of CIoU. $Min$ is the smallest value/index selected. $I_j$ is the index of ground truth corresponding to the minimum cost. We eliminated other matched ground truths according to $I_j$.

\section{Experiments}

\label{exp}

In the interests of fairness, our experimental setup was completely consistent with YOLOv5\cite{yolo5}. In the training process, the same backbone network (CSP\cite{csp}), head network\cite{yolo4}, hyperparameters\footnote{https://github.com/ultralytics/yolov5/blob/master/data/hyps/hyp.scratch-high.yaml}, data augmentation (mosaic\cite{yolo4}, mixup\cite{mixup}, copy-paste\cite{copypaste}), parameter initialization, set random seed, SiLU activation, cosine annealing scheduler\cite{sgdr}, SPP\cite{spp} and mixed-precision training\cite{mpt} were used. The inference procedure was also consistent with YOLOv5, involving NMS, rectangular inference\cite{yolo5}, and half-precision inference. Our experiment used the MS-COCO 2017 dataset\cite{mscoco}(MS-COCO 2017 is licensed under a Creative Commons Attribution 4.0 License) to verify YOLOSA, took YOLOv5 experimental data as the baseline, and used a single graphic card (RTX3090 or RTX A6000) to train only 300 epochs.

\subsection{Ablation Experiment}
\label{exp_ae}

Our ablation experiment mainly compared the innovations with baseline (YOLOv5\cite{yolo5}) data on the MS-COCO VAL2017 dataset\cite{mscoco}. Based on the baseline model(YOLOv5 is licensed under a GNU General Public License v3.0), we added innovations and trained 300 epochs, and then verified the trained model. To ensure the contrast effect of each innovation in the ablation experiment, we only added one innovation to the YOLOv5l model at a time, and ensure that other conditions remained unchanged except innovations, such as hyperparameters, set random seeds, data augmentation, cosine annealing scheduler\cite{sgdr}, mixed accuracy training\cite{mpt}, etc. As summarized in Table ~\ref{table: ae}, each innovation achieved a certain improvement in AP(\%). Among them, our 2D local feature superimposed self-attention (LFSa) module engendered an increase of 0.7\% in AP, the highest increase among all innovations. The efficient decoupled head and AB-OTA also increased AP by 0.4\% and 0.3\% respectively. Therefore, we can conclude that our three innovations are novel and effective.

Due to the limited time and resources, we only added each innovation to the YOLOv5l model for the ablation experiment and did not try to add two innovations at a time. This is not covered by our ablation experiment, but we need to point it out in our paper.

\setlength{\tabcolsep}{4pt}
\begin{table}
\begin{center}
\caption{Ablation Experiment on MS-COCO val2017\cite{mscoco}}
\label{table: ae}
\begin{tabular}{ccccllllll}
\hline\noalign{\smallskip}
Baseline & LFSa & EDH & AB-OTA & AP(\%) & $AP_{50}$ & $AP_{75}$ & $AP_S$ & $AP_M$ & $AP_L$ \\
\noalign{\smallskip}
\hline
\noalign{\smallskip}
\checkmark   &   &   &   & 48.2 & 66.9 & - & - & - & - \\
\checkmark   & \checkmark  &   &   & 48.9(\textcolor{green}{+0.7}) & 67.1(\textcolor{green}{+0.2}) & 53.0 & 31.8 & 53.9 & 62.6 \\
\checkmark   &    & \checkmark &   & 48.6(\textcolor{green}{+0.4}) & 67.2(\textcolor{green}{+0.3}) & 52.8 & 31.2 & 54.0 & 61.6 \\  
\checkmark   &    &   & \checkmark & 48.5(\textcolor{green}{+0.3}) & 67.1(\textcolor{green}{+0.2}) & 52.7 & 32.6 & 53.7 & 62.2 \\
\hline
\end{tabular}
\end{center}
\end{table}
\setlength{\tabcolsep}{1.4pt}

\subsection{Comparative Experiment}

Based on the model size of YOLOv5\cite{yolo5}, our model is also divided into large-scale, medium-scale, and small-scale models. To be fair, the scaling parameters of our model ($depth\_multiple$ and $width\_multiple$) were the same as YOLOv5. As summarized in Table ~\ref{table: ce}, our models with diverse sizes were compared with YOLOv5 models with diverse sizes in terms of AP, parameters, GFLOPs, and latency on the MS-COCO 2017 test-dev dataset\cite{mscoco}. The YOLOSA-S model performs best, exceeding the AP of YOLOv5-S by 3.1\%, and the inference speed is very close to YOLOv5. The AP of YOLOSA-M and YOLOSA-L models also exceeded the YOLOv5-M\cite{yolo5} and YOLOv5-L\cite{yolo5} models by 1.6\% and 0.8\% respectively.

\setlength{\tabcolsep}{4pt}
\begin{table}
\begin{center}
\caption{Comparison of YOLOSA and YOLOv5\cite{yolo5} in terms of AP on MS-COCO 2017 test-dev\cite{mscoco}. All the YOLOSA models are tested at 640*640 resolution, with FP16-precision and batch=1 on a RTX3090.}
\label{table: ce}
\begin{tabular}{lllll}
\hline\noalign{\smallskip}
Models & AP(\%) & Parameters(M) & GFLOPs & Latency(ms) \\
\noalign{\smallskip}
\hline
\noalign{\smallskip}
YOLOv5-S & 36.7 & 7.3 & 17.1 & 8.7 \\
YOLOSA-S & \textbf{39.8}(\textcolor{green}{+3.1}) & 8.9 & 26.1 & 9.3 \\
\hline
YOLOv5-M & 44.5 & 21.4 & 51.4 & 11.1 \\
YOLOSA-M & \textbf{46.1}(\textcolor{green}{+1.6}) & 25.0 & 70.5 & 11.7 \\
\hline
YOLOv5-L & 48.2 & 47.1 & 115.6 & 13.7 \\
YOLOSA-L & \textbf{49.0}(\textcolor{green}{+0.8}) & 53.4 & 147.3 & 14.0 \\
\hline
\end{tabular}
\end{center}
\end{table}
\setlength{\tabcolsep}{1.4pt}

We also compared our models of diverse sizes with YOLOX\cite{yolox} models of diverse sizes. As our baseline is YOLOv5\cite{yolo5}, we cannot be consistent with YOLOX in terms of hyperparameters, data augmentation, parameter initialization, activation functions, and so on. However, we hoped that through comparative experiments, we could understand the advantages of the anchor-based model compared with the anchor-free model and the space for optimization. As listed in Table ~\ref{table: ce2}, YOLOSA-S exceeded YOLOX-S\cite{yolox} by 0.2\% in AP, but the number of calculations and parameters of the model are less than YOLOX. As YOLOSA is an anchor-based model, it will have three times more parameters and calculations than YOLOX in the detection head (each detection head has three anchors). Therefore, we believe that YOLOSA-S is more effective than YOLOX-S. As we did not have a Tesla v100 graphics card and had only one RTX3090 graphics card for inference, we could not make a clear comparison with regard to speed.

\setlength{\tabcolsep}{4pt}
\begin{table}
\begin{center}
\caption{Comparison of YOLOSA and YOLOX\cite{yolox} in terms of AP on MS-COCO 2017 test-dev\cite{mscoco}. All the YOLOSA models are tested at 640*640 resolution, with FP16-precision and batch=1 on a RTX3090.}
\label{table: ce2}
\begin{tabular}{lllll}
\hline\noalign{\smallskip}
Models & AP(\%) & Parameters(M) & GFLOPs & Latency(ms) \\
\noalign{\smallskip}
\hline
\noalign{\smallskip}
YOLOX-S & 39.6 & 9.0 & 26.8 & 9.8 \\
YOLOSA-S & \textbf{39.8}(\textcolor{green}{+0.2}) & 8.9 & 26.1 & 9.3 \\
\hline
YOLOX-M & 46.4 & 25.3 & 73.8 & 12.3 \\
YOLOSA-M & 46.1 & 25.0 & 70.5 & 11.7 \\
\hline
YOLOX-L & 50.0 & 54.2 & 155.6 & 14.5 \\
YOLOSA-L & 49.0 & 53.4 & 147.3 & 14.0 \\
\hline
\end{tabular}
\end{center}
\end{table}
\setlength{\tabcolsep}{1.4pt}

Finally, by comparing the experimental data, we believe that YOLOSA is better than YOLOv5, and surpasses YOLOX in terms of the small-scale model.

\subsection{Comparison with SOTA}

\label{exp_cs}

As presented in Table ~\ref{table: sota}, we compared our approach with other SOTA data. As we did not have a Tesla V100 graphics card and the calculation speed of the model depends on hardware and software and cannot be controlled, we used our existing RTX3090 for FPS calculations. In the interests of fairness, we calculated the FPS by ignoring the time required for post-processing and NMS. After comparing our model with other SOTA approaches, it was found that our models of different sizes exceeded the AP(\%) of YOLOv5\cite{yolo5} when FPS was close to YOLOv5. 

Due to the limitation of time and resources, we did not experiment with smaller size models and larger size models and compared them with other SOTA models. For example, YOLOv5-X\cite{yolo5}, YOLOX-X\cite{yolox}, YOLOX-tiny\cite{yolox}, YOLOv4-tiny\cite{yolo4}, etc. Although some larger size models have obtained higher AP. In addition, we only used a single dataset MSCOCO\cite{mscoco} for the experiment and did not compare it with other SOTA models under more datasets. Because YOLOSA is an anchor-based object detection model, we had not compared it with YOLOX, an anchor-free object detection model. They are not mentioned in section ~\ref{exp_cs}, but they should be within the scope of our paper.

\setlength{\tabcolsep}{1pt}
\begin{table}
\begin{center}
\caption{Comparison of the speed and accuracy of different object detectors on MS-COCO 2017 test-dev\cite{mscoco}. We select all the models
trained on 300 epochs for fair comparison.}
\label{table: sota}
\begin{tabular}{llcccccccc}
\hline\noalign{\smallskip}
Method & Backbone & Size & FPS & AP(\%) & $AP_{50}$ & $AP_{75}$ & $AP_S$ & $AP_M$ & $AP_L$ \\
\noalign{\smallskip}
\hline
\hline
\noalign{\smallskip}
RetinaNet\cite{focal}  & ResNet-50 & 640 & 37 &  37 & - & - & - & - & - \\
RetinaNet\cite{focal}  & ResNet-101 & 640 & 29.4 &  37.9 & - & - & - & - & - \\
RetinaNet\cite{focal}  & ResNet-50 & 1024 & 19.6 &  40.1 & - & - & - & - & - \\
RetinaNet\cite{focal}  & ResNet-101 & 1024 & 15.4 &  41.1 & - & - & - & - & - \\
\hline
YOLOv3 + ASFF*\cite{asff} & Darknet-53 & 320 & 60 &  38.1 & 57.4 & 42.1 & 16.1 & 41.6 & 53.6 \\
YOLOv3 + ASFF*\cite{asff} & Darknet-53 & 416 & 54 &  40.6 & 60.6 & 45.1 & 20.3 & 44.2 & 54.1 \\
YOLOv3 + ASFF*\cite{asff} & Darknet-53 & 608× & 45.5 &  42.4 & 63.0 & 47.4 & 25.5 & 45.7 & 52.3 \\
YOLOv3 + ASFF*\cite{asff} & Darknet-53 & 800× & 29.4 &  43.9 & 64.1 & 49.2 & 27.0 & 46.6 & 53.4 \\
\hline
EfficientDet-D0\cite{efficientdet} & Efficient-B0 & 512 & 62.5 & 33.8 & 52.2 & 35.8 & 12.0 & 38.3 & 51.2 \\
EfficientDet-D1\cite{efficientdet} & Efficient-B1 & 640 & 50.0 & 39.6 & 58.6 & 42.3 & 17.9 & 44.3 & 56.0 \\
EfficientDet-D2\cite{efficientdet} & Efficient-B2 & 768 & 41.7 & 43.0 & 62.3 & 46.2 & 22.5 & 47.0 & 58.4 \\
EfficientDet-D3\cite{efficientdet} & Efficient-B3 & 896 & 23.8 & 45.8 & 65.0 & 49.3 & 26.6 & 49.4 & 59.8 \\
\hline
YOLOv4\cite{yolo4} & CSPDarknet-53 & 608 & 62.0 & 43.5 & 65.7 & 47.3 & 26.7 & 46.7 & 53.3 \\
YOLOv4-CSP\cite{syolo4} & Modified CSP & 640 & 73.0 & 47.5 & 66.2 & 51.7 & 28.2 & 51.2 & 59.8 \\
\hline
YOLOv3-ultralytics\footnotemark[2] & Darknet-53 & 640 & 95.2 & 44.3 & 64.6 & - & - & - & - \\
YOLOv5-S\cite{yolo5} & Modified CSP v5 & 640 & 114.9 & 36.7 & 55.4 & - & - & - & - \\
YOLOv5-M\cite{yolo5} & Modified CSP v5 & 640 & 90.1 & 44.5 & 63.1 & - & - & - & - \\
YOLOv5-L\cite{yolo5} & Modified CSP v5 & 640 & 73.0 & 48.2 & 66.9 & - & - & - & - \\
\hline
YOLOSA-S & Modified CSP v5 & 640 & 107.5 & \textbf{39.8} & 58.1 & 43.0 & 21.2 & 43.4 & 50.7 \\
YOLOSA-M & Modified CSP v5 & 640 & 85.5 & \textbf{46.1} & 64.5 & 50.1 & 26.7 & 50.5 & 57.8 \\
YOLOSA-L & Modified CSP v5 & 640 & 67.6 & \textbf{49.0} & 67.6 & 53.1 & 32.7 & 54.0 & 62.7 \\
\hline
\end{tabular}
\end{center}
\end{table}
\setlength{\tabcolsep}{1.4pt}

\footnotetext[2]{https://github.com/ultralytics/yolov3}

\section{Conclusion}

In this study, we proposed an efficient self-attention module called 2D local feature superimposed self-attention, as well as Efficient Decoupled Head and AB-OTA. These three innovations constitute our model YOLOSA based on the YOLOv5\cite{yolo5} model. YOLOSA achieved very good performance on AP and FPS. On the MS-COCO 2017 dataset\cite{mscoco}, YOLOSA obtained APs of 39.8\%, 46.1\%, and 49.0\%, which exceeded YOLOv5 model’s APs of 3.1\%, 1.6\%, and 0.8\% respectively. Thus, YOLOSA is effective and novel. We hope that our methods can be used in applications in various industries and can help enterprises in different industries reduce costs and improve efficiency while helping them solve the problems of recruitment difficulties and the high cost of employee induction training.

\newpage

{
\small

\bibliographystyle{unsrt}

\begin{thebibliography}{10}

\bibitem{yolo4}
Alexey Bochkovskiy, Chien-Yao Wang, and Hong-Yuan~Mark Liao.
\newblock Yolov4: Optimal speed and accuracy of object detection.
\newblock {\em arXiv preprint arXiv:2004.10934}, 2020.

\bibitem{cascadercnn}
Zhaowei Cai and Nuno Vasconcelos.
\newblock Cascade r-cnn: Delving into high quality object detection.
\newblock In {\em Proceedings of the IEEE conference on computer vision and
  pattern recognition}, pages 6154--6162, 2018.

\bibitem{yolof}
Qiang Chen, Yingming Wang, Tong Yang, Xiangyu Zhang, Jian Cheng, and Jian Sun.
\newblock You only look one-level feature.
\newblock In {\em Proceedings of the IEEE/CVF conference on computer vision and
  pattern recognition}, pages 13039--13048, 2021.

\bibitem{rfcn}
Jifeng Dai, Yi~Li, Kaiming He, and Jian Sun.
\newblock R-fcn: Object detection via region-based fully convolutional
  networks.
\newblock {\em Advances in neural information processing systems}, 29, 2016.

\bibitem{vit}
Alexey Dosovitskiy, Lucas Beyer, Alexander Kolesnikov, Dirk Weissenborn,
  Xiaohua Zhai, Thomas Unterthiner, Mostafa Dehghani, Matthias Minderer, Georg
  Heigold, Sylvain Gelly, et~al.
\newblock An image is worth 16x16 words: Transformers for image recognition at
  scale.
\newblock {\em arXiv preprint arXiv:2010.11929}, 2020.

\bibitem{yolo5}
Glenn~Jocher et~al.
\newblock Yolov5.
\newblock \url{https://github.com/ultralytics/yolov5}, 2021.

\bibitem{yolox}
Zheng Ge, Songtao Liu, Feng Wang, Zeming Li, and Jian Sun.
\newblock Yolox: Exceeding yolo series in 2021.
\newblock {\em arXiv preprint arXiv:2107.08430}, 2021.

\bibitem{copypaste}
Golnaz Ghiasi, Yin Cui, Aravind Srinivas, Rui Qian, Tsung-Yi Lin, Ekin~D Cubuk,
  Quoc~V Le, and Barret Zoph.
\newblock Simple copy-paste is a strong data augmentation method for instance
  segmentation.
\newblock In {\em Proceedings of the IEEE/CVF Conference on Computer Vision and
  Pattern Recognition}, pages 2918--2928, 2021.

\bibitem{fastrcnn}
Ross Girshick.
\newblock Fast r-cnn.
\newblock In {\em Proceedings of the IEEE international conference on computer
  vision}, pages 1440--1448, 2015.

\bibitem{rfh}
Ross Girshick, Jeff Donahue, Trevor Darrell, and Jitendra Malik.
\newblock Rich feature hierarchies for accurate object detection and semantic
  segmentation.
\newblock In {\em Proceedings of the IEEE conference on computer vision and
  pattern recognition}, pages 580--587, 2014.

\bibitem{maskrcnn}
Kaiming He, Georgia Gkioxari, Piotr Doll{\'a}r, and Ross Girshick.
\newblock Mask r-cnn.
\newblock In {\em Proceedings of the IEEE international conference on computer
  vision}, pages 2961--2969, 2017.

\bibitem{spp}
Kaiming He, Xiangyu Zhang, Shaoqing Ren, and Jian Sun.
\newblock Spatial pyramid pooling in deep convolutional networks for visual
  recognition.
\newblock {\em IEEE transactions on pattern analysis and machine intelligence},
  37(9):1904--1916, 2015.

\bibitem{mobilenets}
Andrew~G Howard, Menglong Zhu, Bo~Chen, Dmitry Kalenichenko, Weijun Wang,
  Tobias Weyand, Marco Andreetto, and Hartwig Adam.
\newblock Mobilenets: Efficient convolutional neural networks for mobile vision
  applications.
\newblock {\em arXiv preprint arXiv:1704.04861}, 2017.

\bibitem{senet}
Jie Hu, Li~Shen, and Gang Sun.
\newblock Squeeze-and-excitation networks.
\newblock In {\em Proceedings of the IEEE conference on computer vision and
  pattern recognition}, pages 7132--7141, 2018.

\bibitem{yolopp2}
Xin Huang, Xinxin Wang, Wenyu Lv, Xiaying Bai, Xiang Long, Kaipeng Deng,
  Qingqing Dang, Shumin Han, Qiwen Liu, Xiaoguang Hu, et~al.
\newblock Pp-yolov2: A practical object detector.
\newblock {\em arXiv preprint arXiv:2104.10419}, 2021.

\bibitem{la1}
Kang Kim and Hee~Seok Lee.
\newblock Probabilistic anchor assignment with iou prediction for object
  detection.
\newblock In {\em European Conference on Computer Vision}, pages 355--371.
  Springer, 2020.

\bibitem{fpn}
Seung-Wook Kim, Hyong-Keun Kook, Jee-Young Sun, Mun-Cheon Kang, and Sung-Jea
  Ko.
\newblock Parallel feature pyramid network for object detection.
\newblock In {\em Proceedings of the European Conference on Computer Vision
  (ECCV)}, pages 234--250, 2018.

\bibitem{focal}
Tsung-Yi Lin, Priya Goyal, Ross Girshick, Kaiming He, and Piotr Doll{\'a}r.
\newblock Focal loss for dense object detection.
\newblock In {\em Proceedings of the IEEE international conference on computer
  vision}, pages 2980--2988, 2017.

\bibitem{mscoco}
Tsung-Yi Lin, Michael Maire, Serge Belongie, James Hays, Pietro Perona, Deva
  Ramanan, Piotr Doll{\'a}r, and C~Lawrence Zitnick.
\newblock Microsoft coco: Common objects in context.
\newblock In {\em European conference on computer vision}, pages 740--755.
  Springer, 2014.

\bibitem{pan}
Shu Liu, Lu~Qi, Haifang Qin, Jianping Shi, and Jiaya Jia.
\newblock Path aggregation network for instance segmentation.
\newblock In {\em Proceedings of the IEEE conference on computer vision and
  pattern recognition}, pages 8759--8768, 2018.

\bibitem{asff}
Songtao Liu, Di~Huang, and Yunhong Wang.
\newblock Learning spatial fusion for single-shot object detection.
\newblock {\em arXiv preprint arXiv:1911.09516}, 2019.

\bibitem{ssd}
Wei Liu, Dragomir Anguelov, Dumitru Erhan, Christian Szegedy, Scott Reed,
  Cheng-Yang Fu, and Alexander~C Berg.
\newblock Ssd: Single shot multibox detector.
\newblock In {\em European conference on computer vision}, pages 21--37.
  Springer, 2016.

\bibitem{yolopp}
Xiang Long, Kaipeng Deng, Guanzhong Wang, Yang Zhang, Qingqing Dang, Yuan Gao,
  Hui Shen, Jianguo Ren, Shumin Han, Errui Ding, et~al.
\newblock Pp-yolo: An effective and efficient implementation of object
  detector.
\newblock {\em arXiv preprint arXiv:2007.12099}, 2020.

\bibitem{sgdr}
Ilya Loshchilov and Frank Hutter.
\newblock Sgdr: Stochastic gradient descent with warm restarts.
\newblock {\em arXiv preprint arXiv:1608.03983}, 2016.

\bibitem{la4}
Yuchen Ma, Songtao Liu, Zeming Li, and Jian Sun.
\newblock Iqdet: Instance-wise quality distribution sampling for object
  detection.
\newblock In {\em Proceedings of the IEEE/CVF Conference on Computer Vision and
  Pattern Recognition}, pages 1717--1725, 2021.

\bibitem{mpt}
Paulius Micikevicius, Sharan Narang, Jonah Alben, Gregory Diamos, Erich Elsen,
  David Garcia, Boris Ginsburg, Michael Houston, Oleksii Kuchaiev, Ganesh
  Venkatesh, et~al.
\newblock Mixed precision training.
\newblock {\em arXiv preprint arXiv:1710.03740}, 2017.

\bibitem{eigen}
Mohammed~Bany Muhammad and Mohammed Yeasin.
\newblock Eigen-cam: Class activation map using principal components.
\newblock In {\em 2020 International Joint Conference on Neural Networks
  (IJCNN)}, pages 1--7. IEEE, 2020.

\bibitem{lrcnn}
Jiangmiao Pang, Kai Chen, Jianping Shi, Huajun Feng, Wanli Ouyang, and Dahua
  Lin.
\newblock Libra r-cnn: Towards balanced learning for object detection.
\newblock In {\em Proceedings of the IEEE/CVF conference on computer vision and
  pattern recognition}, pages 821--830, 2019.

\bibitem{yolo1}
Joseph Redmon, Santosh Divvala, Ross Girshick, and Ali Farhadi.
\newblock You only look once: Unified, real-time object detection.
\newblock In {\em Proceedings of the IEEE conference on computer vision and
  pattern recognition}, pages 779--788, 2016.

\bibitem{yolo2}
Joseph Redmon and Ali Farhadi.
\newblock Yolo9000: better, faster, stronger.
\newblock In {\em Proceedings of the IEEE conference on computer vision and
  pattern recognition}, pages 7263--7271, 2017.

\bibitem{yolo3}
Joseph Redmon and Ali Farhadi.
\newblock Yolov3: An incremental improvement.
\newblock {\em arXiv preprint arXiv:1804.02767}, 2018.

\bibitem{fasterrcnn}
Shaoqing Ren, Kaiming He, Ross Girshick, and Jian Sun.
\newblock Faster r-cnn: Towards real-time object detection with region proposal
  networks.
\newblock {\em Advances in neural information processing systems}, 28, 2015.

\bibitem{efficientdet}
Mingxing Tan, Ruoming Pang, and Quoc~V Le.
\newblock Efficientdet: Scalable and efficient object detection.
\newblock In {\em Proceedings of the IEEE/CVF conference on computer vision and
  pattern recognition}, pages 10781--10790, 2020.

\bibitem{selfattention}
Ashish Vaswani, Noam Shazeer, Niki Parmar, Jakob Uszkoreit, Llion Jones,
  Aidan~N Gomez, {\L}ukasz Kaiser, and Illia Polosukhin.
\newblock Attention is all you need.
\newblock {\em Advances in neural information processing systems}, 30, 2017.

\bibitem{syolo4}
Chien-Yao Wang, Alexey Bochkovskiy, and Hong-Yuan~Mark Liao.
\newblock Scaled-yolov4: Scaling cross stage partial network.
\newblock In {\em Proceedings of the IEEE/cvf conference on computer vision and
  pattern recognition}, pages 13029--13038, 2021.

\bibitem{csp}
Chien-Yao Wang, Hong-Yuan~Mark Liao, Yueh-Hua Wu, Ping-Yang Chen, Jun-Wei
  Hsieh, and I-Hau Yeh.
\newblock Cspnet: A new backbone that can enhance learning capability of cnn.
\newblock In {\em Proceedings of the IEEE/CVF conference on computer vision and
  pattern recognition workshops}, pages 390--391, 2020.

\bibitem{cbam}
Sanghyun Woo, Jongchan Park, Joon-Young Lee, and In~So Kweon.
\newblock Cbam: Convolutional block attention module.
\newblock In {\em Proceedings of the European conference on computer vision
  (ECCV)}, pages 3--19, 2018.

\bibitem{mixup}
Hongyi Zhang, Moustapha Cisse, Yann~N Dauphin, and David Lopez-Paz.
\newblock mixup: Beyond empirical risk minimization.
\newblock {\em arXiv preprint arXiv:1710.09412}, 2017.

\bibitem{la3}
Shifeng Zhang, Cheng Chi, Yongqiang Yao, Zhen Lei, and Stan~Z Li.
\newblock Bridging the gap between anchor-based and anchor-free detection via
  adaptive training sample selection.
\newblock In {\em Proceedings of the IEEE/CVF conference on computer vision and
  pattern recognition}, pages 9759--9768, 2020.

\bibitem{la5}
Xiaosong Zhang, Fang Wan, Chang Liu, Rongrong Ji, and Qixiang Ye.
\newblock Freeanchor: Learning to match anchors for visual object detection.
\newblock {\em Advances in neural information processing systems}, 32, 2019.

\bibitem{ciou}
Zhaohui Zheng, Ping Wang, Wei Liu, Jinze Li, Rongguang Ye, and Dongwei Ren.
\newblock Distance-iou loss: Faster and better learning for bounding box
  regression.
\newblock In {\em Proceedings of the AAAI Conference on Artificial
  Intelligence}, volume~34, pages 12993--13000, 2020.

\bibitem{la2}
Benjin Zhu, Jianfeng Wang, Zhengkai Jiang, Fuhang Zong, Songtao Liu, Zeming Li,
  and Jian Sun.
\newblock Autoassign: Differentiable label assignment for dense object
  detection.
\newblock {\em arXiv preprint arXiv:2007.03496}, 2020.

\end{thebibliography}

}

\end{document}